\documentclass[sigconf,nonacm]{acmart}

\AtBeginDocument{%
  }

\usepackage{titlesec}

\definecolor{mygreen}{RGB}{0,150,0}
\definecolor{myred}{RGB}{255,0,0}
\usepackage{enumitem}

\usepackage[font=small,labelfont=bf]{caption}
\usepackage{array,ragged2e}
\usepackage{multicol}
\newcolumntype{P}[1]{>{\RaggedRight\arraybackslash}p{#1}}

\definecolor{paired-light-blue}{RGB}{198, 219, 239}
\definecolor{paired-dark-blue}{RGB}{49, 130, 188}
\definecolor{paired-light-orange}{RGB}{251, 208, 162}
\definecolor{paired-dark-orange}{RGB}{230, 85, 12}
\definecolor{paired-light-green}{RGB}{199, 233, 193}
\definecolor{paired-dark-green}{RGB}{49, 163, 83}
\definecolor{paired-light-purple}{RGB}{218, 218, 235}
\definecolor{paired-dark-purple}{RGB}{117, 107, 176}
\definecolor{paired-light-gray}{RGB}{217, 217, 217}
\definecolor{paired-dark-gray}{RGB}{99, 99, 99}
\definecolor{paired-light-pink}{RGB}{222, 158, 214}
\definecolor{paired-dark-pink}{RGB}{123, 65, 115}
\definecolor{paired-light-red}{RGB}{231, 150, 156}
\definecolor{paired-dark-red}{RGB}{131, 60, 56}
\definecolor{paired-light-yellow}{RGB}{231, 204, 149}
\definecolor{paired-dark-yellow}{RGB}{141, 109, 49}

\definecolor{bg1}{HTML}{FF9966}
\definecolor{bg2}{HTML}{CCE5FF}
\definecolor{bg3}{HTML}{FFCC99}
\definecolor{bg4}{HTML}{FFC107}
\definecolor{bg5}{HTML}{FFCCCC}
\definecolor{bg6}{HTML}{D5E8D4}
\definecolor{bg7}{HTML}{eeeeee}
\definecolor{bg8}{HTML}{cdeb8b}
\definecolor{bg9}{HTML}{dae8fc}
\definecolor{bg10}{HTML}{a2e6eb}

\definecolor{bg31}{HTML}{FFCDD2} 

\definecolor{bg32}{HTML}{F8BBD0}

\definecolor{bg33}{HTML}{E1BEE7} 

\definecolor{bg34}{HTML}{D7CCC8} 

\definecolor{bg35}{HTML}{B2DFDB} 

\definecolor{bg36}{HTML}{A5D6A7} 

\definecolor{bg37}{HTML}{FFF9C4} 

\definecolor{bg38}{HTML}{FFECB3} 

\definecolor{bg111}{HTML}{CB6843}

\definecolor{bg112}{HTML}{D77C5C}

\definecolor{bg113}{HTML}{E28E6E}
\definecolor{bg114}{HTML}{E89F7D}
\definecolor{bg115}{HTML}{EDAE8A}
\definecolor{bg116}{HTML}{F0BA95}
\definecolor{bg117}{HTML}{F3C29F}
\definecolor{bg118}{HTML}{F6CCAA}
\definecolor{bg119}{HTML}{F8D5B3}
\definecolor{bg120}{HTML}{FADCBD}
\definecolor{bg121}{HTML}{FCE6C7}

\definecolor{bg39}{HTML}{FFE0B2} 

\definecolor{bg40}{HTML}{3CB371} 

\definecolor{bg43}{HTML}{ffe5d9}

\definecolor{bg15}{HTML}{7FFFD4}

\definecolor{bg17}{HTML}{F0FFFF}

\definecolor{bg18}{HTML}{F5FFFA}

\definecolor{bg19}{HTML}{F8F8FF}

\definecolor{bg20}{HTML}{FFFFFF}

\definecolor{bg21}{HTML}{E1F5FE}

\definecolor{bg22}{HTML}{B3E5FC}

\definecolor{bg23}{HTML}{81D4FA}

\definecolor{bg24}{HTML}{4FC3F7}

\definecolor{bg25}{HTML}{29B6F6}

\definecolor{bg26}{HTML}{03A9F4}
\definecolor{bg27}{HTML}{039BE5}
\definecolor{bg28}{HTML}{0288D1}
\definecolor{bg29}{HTML}{0277BD}
\definecolor{bg30}{HTML}{01579B}
\definecolor{bg16}{HTML}{FFCC99} 
\definecolor{pg51}{HTML}{E8F5E9} 
\definecolor{pg52}{HTML}{C8E6C9} 
\definecolor{pg53}{HTML}{B9F6CA} 
\definecolor{pg54}{HTML}{A9DFBF} 
\definecolor{pg55}{HTML}{BCF5A6} 

\definecolor{pg56}{HTML}{BEF1CE} 
\definecolor{pg57}{HTML}{CEF6EC} 
\definecolor{pg58}{HTML}{B7F0B1} 
\definecolor{pg59}{HTML}{B1F2B5} 
\definecolor{pg60}{HTML}{9DF3C4} 

\definecolor{pg61}{HTML}{DEF7E0} 
\definecolor{pg62}{HTML}{E8F8DC} 

\definecolor{pg63}{HTML}{EBF7E7} 
\definecolor{pg64}{HTML}{F0FDF4} 

\definecolor{pg65}{HTML}{F1FEE7} 
\definecolor{pg66}{HTML}{F7FFF6} 
\definecolor{pg67}{HTML}{FCFFE7} 
\definecolor{pg68}{HTML}{F4FFD2} 
\definecolor{pg69}{HTML}{EEFFE2} 
\definecolor{pg70}{HTML}{E3FDF5} 

\definecolor{connect-color}{RGB}{0,0,0}
\definecolor{middle-color}{RGB}{255,255,255}
\definecolor{leaf-color}{RGB}{173,216,230}
\definecolor{line-color}{RGB}{25,25,112}

\usepackage{url}            
\usepackage{booktabs}       
\usepackage{amsfonts}       
\usepackage{nicefrac}       
\usepackage{microtype}      
\usepackage{amsmath}
\usepackage{caption}

\usepackage{multirow}
\usepackage{subcaption}
\usepackage{listings}
\usepackage{tcolorbox}
\usepackage{xspace}
\usepackage{makecell}
\usepackage{soul}               
\setstcolor{red}            

\usepackage{longtable}
\usepackage{tablefootnote}

\usepackage[edges]{forest}
\definecolor{hidden-draw}{RGB}{20,68,106}
\definecolor{hidden-pink}{RGB}{255,245,247}
\definecolor{red}{RGB}{255,0,0}

\definecolor{hidden-draw}{RGB}{0,0,0}
\definecolor{hidden-pink}{RGB}{255,182,193}

\usepackage[T1]{fontenc}

\usepackage[utf8]{inputenc}

\usepackage[draft,textsize=footnotesize,textwidth=15mm]{todonotes}

\usepackage{forest}    

\usepackage{tikz}      

\usepackage{hyperref}  

\tikzset{
    root style/.style={
        draw,
        rounded corners,
        fill=blue!30, 
        align=center,
        font=\bfseries
    },
    child style/.style={
        draw,
        rounded corners,
        fill=green!30, 
        align=center,
        font=\bfseries
    },
    grandchild style/.style={
        draw,
        rounded corners,
        fill=red!30, 
        align=center,
        font=\bfseries
    }
}

\tikzset{
  my-box/.style={
    rectangle,
    draw=hidden-draw,
    rounded corners,
    text opacity=1,
    minimum height=1.5em,
    minimum width=40em,
    inner sep=2pt,
    align=center,
    line width=0.8pt,
  },
  leaf/.style={
    my-box,
    minimum height=1.5em,
    text=black,
    align=center,
    font=\normalsize,
    inner xsep=2pt,
    inner ysep=4pt,
    line width=0.8pt,
  }
}

\begin{document}
\setcopyright{none}

\title{Potential and Perils of Large Language Models as Judges of Unstructured Textual Data}

\author{Rewina Bedemariam$^*$}
\affiliation{%
\institution{Amazon}
\city{Arlington}
\state{Virginia}
\country{USA}}

\author{Natalie Perez$^*$}
\affiliation{%
\institution{Amazon}
\city{Honolulu}
\state{Hawai'i}
\country{USA}}

\author{Sreyoshi Bhaduri}
\affiliation{%
\institution{Amazon}
\city{New York}
\state{New York}
\country{USA}}

\author{Satya Kapoor}
\affiliation{%
\institution{Amazon}
\city{Vancouver}
\state{British Columbia}
\country{Canada}}

\author{Alex Gil}
\affiliation{%
\institution{Amazon}
\city{Arlington}
\state{Virginia}
\country{USA}}

\author{Elizabeth Conjar}
\affiliation{%
\institution{Amazon}
\city{Arlington}
\state{Virginia}
\country{USA}}

\author{Ikkei Itoku}
\affiliation{%
\institution{Amazon}
\city{New York}
\state{New York}
\country{USA}}

\author{David Theil}
\affiliation{%
\institution{Amazon}
\city{Arlington}
\state{Virginia}
\country{USA}}

\author{Aman Chadha}
\affiliation{%
\institution{Amazon Gen AI}
\city{Santa Clara}
\state{California}
\country{USA}}

\author{Naumaan Nayyar}
\affiliation{%
\institution{Amazon}
\city{Seattle}
\state{Washington}
\country{USA}}

\thanks{$^*$Both authors contributed equally to this project}
\thanks{Corresponding author: DrSre@amazon.com}



\begin{abstract}

Rapid advancements in large language models have unlocked remarkable capabilities when it comes to processing and summarizing unstructured text data. This has implications for the analysis of rich, open-ended datasets, such as survey responses, where LLMs hold the promise of efficiently distilling key themes and sentiments. However, as organizations increasingly turn to these powerful AI systems to make sense of textual feedback, a critical question arises, can we trust LLMs to accurately represent the perspectives contained within these text based datasets? While LLMs excel at generating human-like summaries, there is a risk that their outputs may inadvertently diverge from the true substance of the original responses. Discrepancies between the LLM-generated outputs and the actual themes present in the data could lead to flawed decision-making, with far-reaching consequences for organizations. 
This research investigates the effectiveness of LLM-as-judge models to evaluate the thematic alignment of summaries generated by other LLMs. We utilized an Anthropic Claude model to generate thematic summaries from open-ended survey responses, with Amazon's Titan Express, Nova Pro, and Meta's Llama serving as judges. This LLM-as-judge approach was compared to human evaluations using Cohen's kappa, Spearman's rho, and Krippendorff's alpha, validating a scalable alternative to traditional human centric evaluation methods. Our findings reveal that while LLM-as-judge offer a scalable solution comparable to human raters, humans may still excel at detecting subtle, context-specific nuances. 
Our research contributes to the growing body of knowledge on AI assisted text analysis. Further, we provide recommendations for future research, emphasizing the need for careful consideration when generalizing LLM-as-judge models across various contexts and use cases.

\end{abstract}

\maketitle

\begin{table*}
    \centering
    \small
    \setlength{\tabcolsep}{4pt}  
    \begin{tabular*}{\textwidth}{@{\extracolsep{\fill}}lcccccc@{}}
    \toprule
Comparisons & Percentage Agreement & Cohen's Kappa & Spearman's Rho & Krippendorff's Alpha Ordinal & Krippendorff's Alpha Nominal \\
\hline
Human vs Claude 2.1 Ratings &{\color{red}79\%} & {\color{blue}0.41} & {\color{blue}0.62} & {\color{red}0.60} & {\color{red}0.42} \\
\hline
Human vs Titan Express Ratings & 78\% & 0.35 & 0.50 & 0.49 & {\color{blue}0.41} \\
\hline
Human vs Sonnet 3.5 Ratings & 76\% & {\color{red}0.44} & 0.60 & {\color{red}0.60} & 0.35 \\
\hline
Human vs Llama 3.3 70b Ratings & {\color{red}79\%}  & 0.39 & {\color{red}0.63} & {\color{red}0.60} & 0.39 \\
\hline
Human vs Nova Pro & 76\% & 0.34 & 0.57 & 0.57 & 0.35 \\
\hline
Claude 2.1 vs Titan Express Ratings & {\color{red}91\%}  & {\color{red}0.70}  & {\color{red}0.86}  & {\color{red}0.87} & {\color{red}0.70} \\
\hline
Claude 2.1 vs Sonnet 3.5 Ratings & 75\% & 0.35 & 0.43 & 0.41 & 0.35 \\
\hline
Claude 2.1 vs Llama 3.3 70b Ratings & 85\% & 0.50 & 0.65 & 0.65 & 0.50 \\
\hline
Claude 2.1 vs Nova Pro Ratings & 85\% & 0.54 & 0.69 & 0.69 & 0.54 \\
\hline
Titan Express vs Sonnet 3.5 Ratings & 74\% & 0.32 & 0.47 & 0.44 & 0.30 \\
\hline
Titan Express vs Llama 3.3 70b Ratings & 84\% & 0.44 & 0.68 & 0.68 & 0.43 \\
\hline
Titan Express vs Nova Pro Ratings & {\color{blue}87\%} & {\color{blue}0.57} & 0.71 & {\color{blue}0.71}& {\color{blue}0.57} \\
\hline
Sonnet 3.5 vs Llama 3.3 70b Ratings & 76\% & 0.37 & 0.55 & 0.49 & 0.37 \\
\hline
Sonnet 3.5 vs Nova Pro Ratings & 82\% & 0.56 & 0.70 & 0.67 & 0.55 \\
\hline
Nova Pro vs Llama 3.3 70b Ratings & 84\% & 0.47 & {\color{blue}0.72} & {\color{blue}0.71} & 0.47 \\
\hline
        \bottomrule
    \end{tabular*} 
    \caption{Inter-rater Agreement Metrics between Human and LLM Rating. {\color{red}Red}: best; {\color{blue}Blue}: second best. }
    \label{tab:lang_perf_indic1}
\end{table*}

\section{Introduction}
The rapid evolution of Large Language Models (LLMs) has expanded their potential uses, from generating content to assessing it. As organizations increasingly adopt these models, there is a growing need to evaluate the accuracy and alignment of LLM-generated outputs with human perspectives \cite{long2023aligning}. The concept of using LLMs as evaluative judges’ dates back to efforts in natural language processing to improve evaluation metrics\cite{wang2024adaptablereliabletextclassification}. However, these traditional evaluation metrics often fall short when it comes to accurately assessing the nuances of natural language generation tasks \cite{liu2023g}. 

As LLMs increasingly power the analysis of open-ended textual data in organizational settings, ensuring the fairness and accuracy of their outputs becomes crucial. In artificial intelligence, model alignment refers to techniques designed to align LLM behaviors with human values and expectations \cite{shen2023large}. This involves methods like fine-tuning, human feedback, and reinforcement learning to ensure outputs reflect human-like reasoning and decision-making processes \cite{liu2023g, balne2024parameter} Model alignment is critical, especially when models are used as judge models in behavioral research \cite{shen2023large}.

Our research makes a significant contribution by investigating the effectiveness of using LLMs as judges to evaluate the thematic alignment of summaries generated by other LLMs, specifically in the context of organizations using open-text survey responses. Our study is important because it addresses a critical gap in the responsible deployment of AI systems for decision-making processes that directly impact organizational decision-making. 

We employed an Anthropic Claude model to generate thematic summaries from open-ended survey responses and utilized Amazon’s Titan Express and Nova Pro LLMs and Llama as judges to evaluate these summaries. By comparing the LLM-as-judge approach with human evaluations using Cohen's kappa, Spearman’s rho, and Krippendorff’s alpha, we present a scalable alternative to traditional human-centric evaluation methods. This research is particularly relevant to the AI in Talent Management research community, as it explores the potential for AI systems to serve as impartial arbiters of content accuracy and representation, while also highlighting the ethical considerations and potential biases inherent in such approaches. 

Our findings demonstrate that LLM-as-judges can produce results comparable to human raters, offering organizations a more efficient means of validating AI-generated insights. However, we also critically examine the limitations of this approach and provide recommendations for future research to ensure fairness, accountability, and transparency in the use of LLMs for organizational decision-making. This work contributes to the ongoing dialogue on responsible AI deployment and the development of trustworthy evaluation mechanisms for AI-generated content in high-stakes environments. 

The following research questions are explored: 
\begin{enumerate}
\item To what extent can LLMs replicate human judgment in evaluating thematic alignment, and what factors contribute to discrepancies between LLM and human ratings?
\item What are the implications of higher inter-model agreement compared to human-model agreement for the development and application of LLMs in similar content analysis and theme evaluation tasks?
\end{enumerate}


\section{Background}
LLM-as-a-judge has emerged as an innovative solution, wherein an LLM evaluates another model’s output to approximate human labeling \cite{zheng2023judging}. This approach offers the promise of automating human judgment at scale, while maintaining high levels of reliability and consistency. For example, \cite{yan2024evaluating} demonstrated that, when properly calibrated, LLMs can reach agreement rates close to those of human annotators, which can be costly and time-intensive. From this lens, the utilization of LLMs can significantly reduce analysis time and turn around rapid results.  By comparing an LLM's evaluation of generated content to human evaluations, researchers aim to refine LLM capabilities and ensure alignment with human interpretations \cite{rao2024quallm, wang2024adaptablereliabletextclassification}.This approach could simplify the evaluation process by reducing reliance on human resources, enabling faster iterations in development cycles. Therefore, understanding how well LLMs can replicate and adhere to human judgment is crucial for their effective deployment in organizational settings \cite{wang2024adaptablereliabletextclassification}.

\subsection {Open-Text Survey Data}
Open-ended survey data is a type of textual data where respondents provide unstructured text-based responses. This type of data is useful for obtaining authentic and sometimes unexpected information that illuminates “why” or “how” respondents think, feel, or behave \cite{rouder2021all}. Unlike scaled or categorical survey items, open-ended questions encourage respondents to provide information about their experiences, perceptions, thoughts, values, and/or feelings that researchers might not anticipate or share information about topics that might be sensitive or personal in nature \cite{allen2017sage}. Open-text data can be particularly valuable in telling nuanced stories and highlighting diverse responses \cite{rouder2021all}. 

\subsection{Thematic Summaries}
LLMs can be used for a variety of textual analysis, including topic modeling \cite{bhaduri2024reconciling, kapoor2024qualitative} and summarizing large amounts of unstructured text-based data \cite{tang2023evaluating}. The objective of textual summarization is to communicate the primary meaning of an original dataset into a simplified and straightforward form without sacrificing the integrity of the information being conveyed \cite{dutta2022data}. Scholars in the LLM field have explored textual summarization as a valuable use-case for LLMs \cite{tang2023evaluating}. Yet research has revealed factual inconsistencies in using LLMs to generate textual summaries, with a range of errors or biases \cite{tang2023evaluating, ashwin2023using}. 

\subsection{LLMs and Classification Evaluation Strategy}
There are a number of LLM evaluation strategies available, and classification prompts are one way to classify text into predefined categories \cite{rao2024quallm}. In particular, classification prompts allow researchers to classify content into categories, inductively or deductively \cite{rao2024quallm}. Historically, text classification has been conducted using traditional ML approaches that are often complex and resource-intensive; tasks include extraction, dimensionality reduction, classifier selection, and model evaluation, not including pre-processing steps \cite{wang2024adaptablereliabletextclassification}. However, LLM advancements have made text classification easier with only three required elements: 1) data collection, 2) feeding LLM data, and 3) obtaining classification results from an LLM \cite{wang2024adaptablereliabletextclassification}. Since LLMs have been pre-trained on diverse datasets, researchers have found that they require little, if any, additional training specific to a task or domain area \cite{wang2024adaptablereliabletextclassification}. 

Given recent LLM advancements and prompt engineering, LLMs offer a cheaper alternative to classify thematic summaries based on unstructured open-text survey data. This ability has the potential to help researchers not only analyze textual data, but also classify textual generation into distinct categories \cite{zhao2023recommender}. In particular, researchers can use classification prompts and evaluation to assess the accuracy, also known as correctness, of LLM classifications, by measuring the proportion of accurately classified thematic summaries \cite{rao2024quallm}. That said, the absence of robust methodological strategies to evaluate open-text thematic summaries reveals a gap in the research field addressing this novel use-case. This study aims to evaluate how well classification prompts can accurately rate thematic summary content alignment.


\section{Methods}
The process of implementing the LLM-as-judge methodology involves taking the text output from one AI model and feeding it back into another LLM. The second model, now serving as the LLM-as-judge, evaluates the text based on an evaluation prompt provided by the user. The second LLM then returns a score, label, or descriptive judgment depending on the specific evaluation criteria set by the user. This allows for a high degree of customization, as users can instruct the LLM to assess specific properties, which makes the approach adaptable to various applications. This framework also means that LLM-as-judge is not an evaluation metric in the traditional sense, like accuracy or precision \cite{yan2024evaluating}. Rather, it serves as a general technique for approximating human judgment, where the LLM relies on its training to assess qualities like “faithfulness to source” or “correctness.” This way, LLMs act as proxy evaluators, following detailed prompts much like a human evaluator would. While this technique does not produce a fixed measure, it offers a flexible proxy metric that can align with specific use cases.

\subsection{Dataset}
We obtained access to a novel, non-open source dataset for this analysis. The benefit of testing across a new dataset, not previously used for model pre-training, is that the researchers can determine how well LLMs analyze and generate predictions (outputs) accurately. The data used in this study was collected using a census survey that addressed work-related topics and was launched across a multi-global population who reported working full-time. The survey contained one free-text question that asked the respondents to share one thing about work they want senior leaders to be aware of. Over 13,000 comments were collected. The researchers pre-processed the responses by removing short samples, noise (e.g., text with only symbols such as periods or dashes and no other content), and personally identifiable information (PII). The 13K dataset was further grouped into 70 smaller datasets by segmenting the data into groups based on business lines, so insights could be shared with business lines and the datasets would not be corrupted with comments from individuals that did not report to certain business lines. 

Next, a previously engineered thematic summary prompt was used to analyze the dataset and create LLM-generated thematic summaries. The thematic summaries were produced based on a validated prompt that leverages thematic quality principles, as outlined by \cite{naeem2023step}, including JSON formatting instructions focused on identifying a theme name (must contain at least one topic and sentiment details), thematic summary (must contain words that reflect genuine experiences or perceptions of respondents, must contain words that are rich in meaning and provide detailed understanding of the topic or focus area, must describe the relationship of theme to related topics that aid to contribute to new insight), and raw verbatim comments (must include verbatim words expressed by respondents). The data was run through different LLM API calls. A total of 70 thematic summaries containing three themes per summary were generated (i.e., each theme contained a theme name, theme description with 3-4 sentences, and one representative verbatim comment), based on the 70 input datasets. The thematic summaries provide aggregated findings across the top-three most salient and prevalent themes within each of the respective input datasets. This study only used the 70 LLM-generated thematic summaries to evaluate how well the thematic summary content was aligned across theme name, theme description, and representative comments to reflect the theme description and theme name. 

\subsection{Overview of Evaluation}
Our general methodology used a Claude model to generate thematic summary outputs from survey comments. For evaluations, this study utilized a three-stage methodology to evaluate the thematic alignment of LLM-generated summaries, which involved human evaluators, and several LLM models including Anthropic Claude (v2.1), Claude Sonnet (v3.5) Amazon Titan Express and Nova Pro, and Llama evaluations. Each model was tasked with assessing alignment across three dimensions: theme name, description, and representative quote.

\subsubsection{Step 1. Human Evaluation as Baseline.} Human evaluators were first tasked with reviewing the thematic summaries based on alignment among the theme name, description, and representative verbatim comment/quote generated by a Claude model. The human ratings served as the baseline for comparison against LLM assessments. To evaluate the prompt’s ability to accurately evaluate content alignment, human evaluators used the same content alignment criteria as the LLM-as-judge model (see scale below). 
\subsubsection{Step 2. LLM Evaluation: Claude as the Initial Evaluator.} The Claude model was provided with the same summaries and instructed to assign alignment scores based on a structured evaluation prompt. The prompt specifically asked Claude to rate the thematic coherence across each theme's name, description, and quote. The prompt engineered for this study to classify thematic summaries into predefined categories was based on content alignment using the following rating scale: 1 to 3, with 1 (“Not Aligned"), 2 ("Somewhat Aligned"), and 3 ("Completely Aligned"). Since each thematic summary contained three dimensions, the prompt evaluated each dimension independently (refer to Appendix for prompt and output examples).
\subsubsection{Step 3. Multiple Models as LLM-as-judge of Claude’s Output.} Following Claude’s evaluation, Titan Express, Claude Sonnet 3.5, Llama 3.3 (70b), and Nova Pro models were tasked with evaluating the scores generated by the human evaluations and across the models to identify a baseline for agreement cross the human and model evaluations. Inferential parameters were configured with the following settings: top-p value of 0.9, top-k value: 0.25, and a Temperature of 0. This multi-model evaluation aimed to mirror a human review process while leveraging different models' unique capabilities. This experiment compared the evaluation results of several distinct LLM ratings of the same outputs. The objective was to determine if LLMs could replicate each other’s judgments and thus validate the consistency of the LLM-as-judge approach. Claude, Titan, Nova, and Llama models served as the evaluators, examining thematic alignment, accuracy, and relevance of the outputs. 

\section{Evaluation Strategy}
The evaluation strategy for this study involved testing and validating a classification prompt focused on content alignment, then using the prompt and running Anthropic’s Claude on the 70 thematic summaries generated from over 13,000 open-ended survey responses. A blind review was conducted with human evaluators to rate the same 70 thematic summaries. Both the classification prompt and human raters were assigned the same directions and rating criteria to rate the thematic summaries based on content accuracy. Reliability was assessed using four metrics: Percentage Agreement, Cohen’s kappa, Spearman’s rho, and Krippendorff’s alpha. Percentage Agreement provides an intuitive measure of exact matches but does not account for chance alignment. Cohen’s kappa adjusts for chance but assumes nominal categories. Spearman’s rho determines if enough data has been rated, while Krippendorff’s alpha (Ordinal) accounts for ordinal data’s inherent structure, offering a more nuanced perspective on agreement \cite{cohen1960coefficient, warrens2015five, hayes2007answering}.

Evaluation Processes
A structured evaluation process was employed to determine the alignment between the LLM and human ratings. The process involved three key steps:
\begin{enumerate}
    \item Evaluation Classification Prompt Development. A classification prompt was engineered and used by human evaluators through reviewing and rating 70 LLM-generated thematic summary outputs. The aim of the prompt engineering focused on the scope of the evaluation, whether to evaluate the entire summary prompt, or evaluate each theme independently within one summary prompt. The researchers decided to focus on engineering a prompt to focus on each theme independently across each thematic summary, which would result in three ratings per thematic summary, since each thematic summary contained three themes. 
    \item Prompt Testing and Validation. The classification prompt underwent testing to ascertain its effectiveness in identifying misalignment. This phase was conducted in partnership between different research teams, ensuring the prompt was accurate and reliable for broader application in model alignment tasks. The aim of the prompt was to accurately classify content alignment within thematic summaries. Multiple iterations were conducted until the research teams determined the details, instructions, and results aligned with the anticipated outputs, including ratings for each independent theme and details that describe the LLM’s reason(s) for classifying each theme into a particular category (refer to Appendix A for prompts used for the evaluations). 
    \item Implementation. After finalizing the classification prompt, all 70 thematic summaries were run through 70 different LLM API calls. The researchers securely stored the ratings for further comparative analysis with the human evaluations, LLM evaluations and the LLM-as-judge model. 
\end{enumerate}

\section{Results}
Our study sought to answer two research questions: 1) To what extent can LLMs replicate human judgment in evaluating thematic alignment, and what factors contribute to discrepancies between LLM and human ratings? and 2) What are the implications of higher inter-model agreement compared to human-model agreement for the development and application of LLMs in content analysis and theme evaluation tasks? 

We used the following metrics to compare alignment and consequently, performance, across and among models and humans: \begin{enumerate}
\item Percentage Agreement definition: Percentage agreement tells us how many times two raters provide the same rating (e.g., 1 – 5) of the same thing, such as two people providing the same 5-star rating of a movie. The more times they agree, the better. This is expressed as a percentage of the total number of cases rated and calculated by dividing the total agreements by the total number of ratings and multiplying by 100 \cite{mchugh2012interrater}. 

\item Cohen’s kappa definition: Cohen’s kappa is essentially a smarter-version of percentage agreement. It is like when two people guess how many of their 5 co-workers will wear the color blue in the office each day; sometimes both people guess the same number (e.g., 1—5) by chance. Cohen’s kappa takes into account how well the two people agree, beyond any lucky guesses. The coefficients range from -1 to +1, where 1 represents perfect agreement, 0 represents agreement equivalent to chance, and negative values indicate agreement less than chance \cite{cohen1960coefficient}.

\item Spearman’s rho definition: Spearman's rho is like a friendship meter for numbers. It shows how well two sets of numbers “get along” or move together. If one set of numbers goes up and the other set also goes up, they have a positive relationship. If one goes up while the other goes down, they have a negative relationship. Coefficients range from 1 to +1, with values closer to ±1 indicate stronger correlations \cite{schober2018correlation}. In this study, rho was used as a statistical test on Cohen’s kappa as a parameter. Rho allowed researchers to determine if enough data was used to ensure the rating agreement was sound \cite{shaffer2017quantitative}. 

\item Krippendorff's alpha definition: Krippendorff's alpha is a test used to determine how much all raters agree on something. Imagine two people taste-testing different foods at a restaurant and rating the foods on a scale of 1—5. Krippendorff's alpha provides a score to show how much the two people agree on their food ratings, even if they did not taste every dish in the restaurant. The alpha coefficient ranges from 0 to 1, where values closer to 1 indicate higher agreement among raters. Generally, an alpha above 0.80 signifies strong agreement, between 0.67 and 0.80 indicates acceptable agreement, and below 0.67 suggests low agreement \cite{krippendorff2011computing}. If calculated with the rationale that the levels (1, 2 and 3) are ordinal, Krippendorff’s Alpha considers not just agreement but also the magnitude of disagreement. It is less affected by marginal distributions compared to kappa and provides a more nuanced assessment when ratings are ranked (ordinal). That is, while percentage agreement and kappa treat all disagreements equally, Alpha recognizes the difference between minor (e.g., "1" vs. "2") and major disagreements (e.g., "1" vs. "3"; \cite{krippendorff2018content}.
\end{enumerate}

We examined the alignment between ratings provided by a LLM and human evaluators for a set of 70 thematic summaries. The inter-rater agreement between human evaluators and LLMs demonstrated consistency across various LLM architectures. When comparing the Cohen’s kappa results between the human ratings and the models, Sonnet 3.5 had the highest rate of agreement with human raters with a score of 0.44, indicating moderate agreement \cite{cohen1960coefficient}. However, nuanced differences emerged when examining Cohen's kappa and Krippendorff's alpha coefficients. This variability in agreement metrics can be attributed to several factors: the differential sensitivity of agreement measures, with Cohen's kappa being more sensitive to marginal distributions and Krippendorff's alpha accounting for different types of data and multiple raters; the inherent stochasticity in LLM outputs, which introduces a degree of randomness potentially affecting the stability of agreement metrics; task-specific performance variations, where certain LLMs may exhibit slight advantages in specific evaluation criteria or content domains; and variations in human raters' expertise and their interpretation of annotation guidelines. These findings underscore the importance of employing multiple agreement metrics and considering the underlying factors that influence inter-rater reliability when assessing LLM performance in evaluation tasks.

\subsection{Moderate Agreement Between LLM and Human Ratings}
Our results show that despite the high percentage agreement between the human ratings and those of the models (e.g., two highest percentage agreements – Human vs. Claude (v2.1): 79\% and Human vs. Llama: 79\%), the corresponding Cohen's kappa and Krippendorff’s alpha values indicate a range of low to high substantial reliability (Kappa range between human and models: 0.34–0.44; Kappa range between models: 0.32—0.70; Alpha range between human and models (ordinal): 0.49—0.60; Alpha range between models (ordinal): 0.41—0.87; Alpha range between human and models (nominal): 0.35–0.42; Alpha range between models: 0.30—0.70). The Spearman’s rho results suggest moderate to strong correlation between the human and model ratings (0.50—0.62). In most cases, Spearman's rho is higher than Cohen's kappa. This suggests that while the humans and models might not always provide the exact same rating, they tend to rank items similarly. In addition, while percentage agreement only considers exact matches, metrics like Cohen’s Kappa and Krippendorff’s alpha adjust for chance agreement and weighting of ordinal disagreements, respectively (refer to Table 1). Krippendorff’s alpha further accounts for the ordinal nature of the data, reflecting nuanced differences between raters more accurately than percentage agreement (Artsein and Poesio, 2008; Krippendorff, 2018).

\subsection{Variability in Performance Across LLM Models}
Interestingly, when considering all the metrics, Claude (v.2.1) performed well across all metrics, with one of the highest percentage agreements with the human ratings (79\%), second-highest Cohen's Kappa score (0.41), tied for highest Krippendorff's alpha (Ordinal) score (0.60), and highest Krippendorff's alpha (Nominal) score (0.42). The model that performed the second best across all metrics was Llama 3.3 (70b). Sonnet 3.5 and Nova Pro did not perform as well across most metrics. These findings suggest that Claude (v2.1) appears to be the most consistently aligned with human ratings across all metrics, followed by Llama 3.3, with Titan Express, Sonnet 3.5, and Nova Pro presenting mixed results. Although the choice of prioritizing metrics could change the ranking, Claude (v2.1) is most consistent across all metrics. As an older model in the Claude suite, the results suggest that “newer” models do not always produce better results for all use-cases. Researchers might benefit from examining several models to determine which bests suites their respective use-cases.  

\subsection{Higher Inter-Model Agreement Compared to Human-Model Agreement}
Overall, the results suggest a moderate to high level of agreement and reliability between the human and LLM model ratings, as well as between the different LLM model ratings. However, when examining the results more closely, the models indicate being generally more consistent with each other compared with the human ratings, as evidenced by the higher percentage agreement, Cohen's Kappa, and Krippendorff's alpha values for the model-to-model comparisons. Overall, our findings suggest that while models such as Claude’s Sonnet 2.1 demonstrates a reasonable degree of alignment with human evaluations, there remains room for improvement. The moderate Cohen's Kappa highlights some discrepancies between human and LLM ratings. It is possible that the observed variations could stem from nuanced differences in human interpretation, especially when themes involve complex issues. Examples of discrepancies included instances where the LLM rated themes as completely aligned despite minor misalignments identified by human evaluators.

Finally, our study results highlighted areas of alignment and discrepancies between LLM and human ratings:

\begin{itemize}
\item High Agreement Cases. In many instances where both the LLM and human rated the themes as “Completely Aligned,” the thematic summaries displayed a high degree of coherence across theme name, description, and quotes. Both raters frequently aligned in recognizing specific work environment topics. Topics that were aligned included but were not limited to topics addressing role function, leadership, and policies.  

\item Discrepancies and Over-Estimation by LLM. In cases where the human rated alignment as “somewhat aligned” or “not aligned,” the LLM sometimes rated itself as “Completely Aligned.” For example, the LLM might have focused on general content similarities, while the human raters noted discrepancies in specific details, such as incomplete coverage of the theme or misaligned quotes. This finding suggests that humans might perform better with more nuanced content compared with LLMs. 
\end{itemize}

\section{Recommendation}
To address the observed discrepancies, researchers should consider incorporating additional evaluation metrics beyond content alignment. Possible improvements include investigating and mitigating biases in the evaluation of LLMs, which is crucial for refining the assessment process and ensuring reliable outcomes. Position bias, which favors options presented earlier, and verbosity bias, which can lead to an overvaluation of longer responses, are among the key challenges that need to be addressed \cite{saito2023verbosity}. Other potential biases include recency bias, confirmation bias, and anchoring bias. Mitigating these biases requires a multifaceted approach, encompassing careful design of evaluation protocols, randomization of response order, utilization of diverse evaluators, and development of objective metrics \cite{saito2023verbosity}. Defining comprehensive success metrics for LLMs necessitates interdisciplinary contributions from various fields \cite{bhaduri2024multi, mackenzie2024beyond}. Computer science and AI can develop specific capability benchmarks and quantify properties like coherence and factual accuracy. Linguistics can assess grammatical correctness and pragmatic aspects of communication. Psychology can design experiments to measure human preferences and evaluate cognitive load. Philosophy can explore ethical considerations and refine definitions of key concepts. Domain experts can assess task-specific performance, while sociology and anthropology can examine societal implications and cultural sensitivities. Human-computer interaction can focus on user experience, and statistics can develop robust methodologies for data analysis. The integration of these diverse perspectives into cohesive evaluation frameworks represents an ongoing challenge in the rapidly evolving field of LLM development and assessment.

\section{Conclusion}
This study contributes to a deeper understanding of how well LLMs align with human judgments in thematic analysis. While 
the percentage agreement and Cohen’s Kappa results indicate fair agreement, the findings point to the need for ongoing adjustments and improvements to the LLM’s evaluation prompt. The analysis reveals areas for improvement, particularly in instances where the LLM overestimated the degree of alignment compared to human raters. This tendency may stem from the LLM's limitations in fully comprehending the nuanced details and contextual factors that contribute to human interpretations of thematic content. The discrepancies highlight the continued need for human oversight and the refinement of evaluation frameworks to better capture the qualitative aspects of thematic understanding.

Future research should explore strategies to further enhance the LLM's alignment with human judgments. This may involve developing more sophisticated prompts and evaluation criteria that account for thematic salience, repetition of quotes, and other contextual factors. Additionally, fine-tuning the LLM to better identify and handle personal identifiable information (PII) and theme recurrence could improve the reliability and trustworthiness of the generated insights. As researchers continue to leverage the power of LLMs in analyzing open-ended survey data, ongoing collaboration between human experts and machine learning systems will be crucial. By iteratively improving the evaluation methods and incorporating human feedback, researchers can unlock the full potential of LLMs to generate high-quality thematic summaries that reliably reflect the perceptions and experiences of survey respondents.

\bibliographystyle{ACM-Reference-Format}
\bibliography{sample-base}
\clearpage

\onecolumn
\appendix
\section{Appendix}
Note: The content in this prompt is mock-content generated to provide an example of the thematic structure only; it does not reflect real data used in this study.

\subsection*{Input Pseudo-Prompt}
\begin{tcolorbox}[colback=gray!5,colframe=gray!40,title=Task Description]

You are a behavioral research scientist whose job is to review outputs for thematic alignment. You will skillfully review, analyze, and determine whether or not the content in the <theme></theme>, <description></description>, and <quote></quote> are aligned in meaning or not using this scale:1 (not aligned), 2 (somewhat aligned), or 3 (completely aligned).  

Follow these steps in order and do NOT to miss theme2 and theme3:
\begin{enumerate}
    \item Review the information in each theme: <theme1></theme1>, <theme2></theme2>, <theme3></theme3>
    \item Check for alignment between the meaning within each of the themes: <theme1></theme1>, <theme2></theme2>, and <them3></theme3>.
    \item Based on each theme reviewed in step 1,2,and 3 assign an alignment rating score for each theme based on the following scale: 1 (not aligned), 2 (somewhat aligned), or 3 (completely aligned).
\end{enumerate}

\textbf{Assistant:}
\begin{itemize}
    \item Rating score for each theme
    \item Reasoning for each rating
\end{itemize}
\end{tcolorbox}

\subsection*{Themes}
\begin{tcolorbox}[colback=gray!5,colframe=gray!40,title=Task Description]
\textit{Theme 1: Learning and Development Desires}
\begin{itemize}
    \item \textbf{Description:} Many individuals expressed their satisfaction with learning and development opportunities. However, some individuals requested more opportunities to engage in multi-day hands-on learning sessions with expert mentors to aid in learning and skill development. Some individuals suggested more learning and development trainings on a more regular cadence.  
    \item \textbf{Quote:} "I really love the opportunities to learn and grow here. There are a lot of options for new and tenured people to learn. That said, I would love more hands-on training options with mentors; it would be great to be able to ask questions with an expert on-hand."
\end{itemize}
\end{tcolorbox}

\begin{tcolorbox}[colback=gray!5,colframe=gray!40,title=Task Description]
\textit{Theme 2: Recognition Perceived as Important}
\begin{itemize}
    \item \textbf{Description:} Individuals indicated there are many different recognition opportunities for career development, training, and promotion, some feel this is contributing to motivation and engagement. The ability to provide pathways for recognition and advancement was frequently cited as something that is valued. Some individuals connected recognition and advancement, reinforcing perceptions of value associated with recognition. 
    \item \textbf{Quote:} "Lot of opportunities to be recognized, which is something I appreciate. Being recognized is important to advancement I think."
\end{itemize}
\end{tcolorbox}

\begin{tcolorbox}[colback=gray!5,colframe=gray!40,title=Task Description]
\textit{Theme 3: Global Connections}
\begin{itemize}
    \item \textbf{Description:} There is a perception that the culture is valued as innovative and there are many opportunities to connect with individuals globally. Having the ability to partner, interact, and work with different individuals from around the globe was valued and meaningful. Some individuals feel opportunities to continue cross-cultural exchange opportunities. 
    \item \textbf{Quote:} I feel it is amazing to be able to collaborate with peers around the world. Would be great to be able to continue to partner in a variety of ways. 
\end{itemize}
\end{tcolorbox}

\subsection*{LLM-as-judge Output Example}
\begin{tcolorbox}[colback=gray!5,colframe=gray!40,title=Alignment Ratings]

\textbf{Theme 1 Rating: 3}\\
\textit{Reason:} The theme, description, and quote in theme1 are completely aligned. They all focus on learning and development opportunities, with the description and quote both mentioning satisfaction with current opportunities while also expressing a desire for more hands-on, mentor-led training sessions.

\vspace{0.5cm}

\textbf{Theme 2 Rating: 3}\\
\textit{Reason:} The theme, description, and quote in theme2 are completely aligned. They all emphasize the importance of recognition in the workplace, its connection to career advancement, and how it contributes to motivation and engagement. The quote directly supports the description by highlighting the value placed on recognition opportunities.

\vspace{0.5cm}

\textbf{Theme 3 Rating: 3}\\
\textit{Reason:} The theme, description, and quote in theme3 are completely aligned. They all focus on the value of global connections and cross-cultural collaborations within the organization. The description and quote both express appreciation for the opportunity to work with individuals from around the world and a desire to continue and expand these global partnerships.
\end{tcolorbox}

\end{document}